# MACHINING STRATEGY CHOICE: PERFORMANCE VIEWER.

**Laurent TAPIE**
LURPA - ENS Cachan, 61 avenue du Président Wilson 94235 Cachan cedex, France
Tél. : (33) 1 47 40 27 64, Fax : (33) 1 47 40 22 20, e-mail : laurent.tapie@lurpa.ens-cachan.fr
**Kwamivi Bernardin MAWUSSI**
LURPA - ENS Cachan, 61 avenue du Président Wilson 94235 Cachan cedex, France
Tél. : (33) 1 47 40 24 13, Fax : (33) 1 47 40 22 20, email: kwamivi.mawussi@lurpa.ens-cachan.fr
IUT de Saint Denis – Université Paris 13, Place du 8 Mai 1945 93206 Saint-Denis cedex, France
**Bernard ANSELMETTI**
LURPA - ENS Cachan, 61 avenue du Président Wilson 94235 Cachan cedex, France
Tél. : (33) 1 47 40 29 71, Fax : (33) 1 47 40 22 20, email: bernard.anselmetti@lurpa.ens-cachan.fr
IUT de Cachan – Université Paris 11, 9 avenue de la division LECLERC 94234 Cachan Cedex, France

## Abstract:

*Nowadays high speed machining (HSM) machine tool combines productivity and part quality. So mould and die maker invested in HSM. Die and mould features are more and more complex shaped. Thus, it is difficult to choose the best machining strategy according to part shape. Geometrical analysis of machining features is not sufficient to make an optimal choice. Some research show that security, technical, functional and economical constrains must be taken into account to elaborate a machining strategy. During complex shape machining, production system limits induce feed rate decreases, thus loss of productivity, in some part areas. In this paper we propose to analyse these areas by estimating tool path quality. First we perform experiments on HSM machine tool to determine trajectory impact on machine tool behaviour. Then, we extract critical criteria and establish models of performance loss. Our work is focused on machine tool kinematical performance and numerical controller unit calculation capacity. We implement these models on Esprit® CAM Software. During machining trajectory creation, critical part areas can be visualised and analysed. Parameters, such as, segment or arc lengths, nature of discontinuities encountered are used to analyse critical part areas. According to this visualisation, process development engineer should validate or modify the trajectory.*

## 1 Introduction

The high quality and high productivity in machining of complex parts has been of primary interest in manufacturing industries these last years. To improve the efficiency of NC machining, high speed machining (HSM) is now widely used in moulds and dies manufacturing industries [1]. The complex shapes of these moulds and dies are rather defined as complex features which are mainly represented by their geometric parameters [2]. The geometric representation of features has been widely explored in several works [3-4]. In conventional machining, the geometric features are associated to machining features through a recognition process [5-6]. In an HSM context, the machining strategies applied, such as Z-level or parallel planes [7], lead to the manufacturing of the whole part in a single sequence [8]. At this stage, the geometric parameters are not sufficient for the development of an efficient machining process. From the results of the machining simulation, A. Dugas [9] showed that it is necessary to take into account security, technical, functional and economical constrains in the development of machining processes. During the machining of the complex features, it is difficult to keep the programmed feed rate in some areas. The feed rate reduction which is observed in these areas is closely linked with the geometric variation of complex features shapes and technical constraints of the machine tool. Several works were devoted to these technical constraints [10] which can be classified as follow [9]: machine tool, numerical controlled unit (NCU) and cutting tool limits.

When analysing the geometric variation of shapes, the technical constraints of the machine tool must be integrated in the development of the machining strategy. This integration is generally difficult because machine tool and NCU limits are not well known. The specialist does not have tools such as monitoring to associate the technical constraints to the geometric variation of the shapes. In this paper, we propose an original procedure to make easier the choice of machining strategies. This procedure is based on experimental observations of our machine tool, tool paths analysis and machining performance visualization. The experimental observations are conducted according to kinematical constrains and NCU calculation capacity. They lead to the modelling of the machine tool behaviour. The tool paths are analysed according to the instantaneous feed rate which is computed. The visualization of the performance is done directly in the work space of the Esprit® CAM software in order to assist the specialist.

## 2 HSM technical constrains

When machining complex features, important variations of the shapes geometrical characteristics lead to a feed rate reduction according to machine tool and NCU limits. This feed rate reduction decreases the part accuracy as the programmed tool path is not respected. The geometrical characteristics are mainly tangential and curvature discontinuities. When these discontinuities appear in a program, the NCU adjusts in an anticipated way the feed rate in order to compensate for its brutal drop: this is the static look ahead phenomena.

To underline the static look ahead phenomena and to elaborate the indicators useful for the analysis of tool paths, we carried out experiments in a milling center MIKRON UCP 710 with SIEMENS controller 840D. Records of the feed rate were carried out thanks to its integrated oscilloscope reading output encoder data. The experimental results are presented afterwards.

### 2.1 Machine tool cinematic

*Tangential discontinuities*

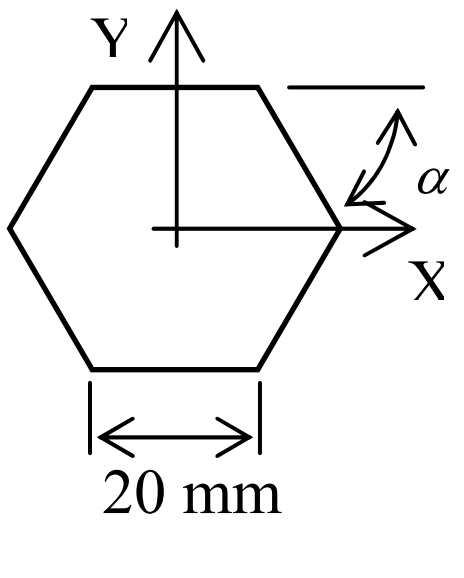

$$\alpha \in [5-10-20-30-40-60-90-120°]$$

$$Feed_speed \in [2-4-8-12m/\min]$$

*Figure 1. Programmed trajectories*

To highlight the machine tool behaviour during tangential discontinuity crossing between two consecutives segments, we carried out the experiments proposed in [11]. The tool path is a polygonal shape and the factors considered are the corner angle α (given by the number of segments in the polygon) and the programmed feed rate (Figure 1). The segments have the same length.

Theoretically to cross a tangential discontinuity with a given feed rate, the machine tool must produce an infinite acceleration [9]. To avoid this impossible increase in acceleration, a solution consists in forcing a null feed rate at the end of each program block. In this case, vibrations can affect the machining accuracy. Another solution is to modify the tool path in order to keep the programmed feed rate. But the modification of the tool path introduces variations which can also affect the machining accuracy. Most of HSM controllers integrate a hybrid solution whereby the programmed feed rate is reduced in order to control the variations induced by the modification of the tool path.

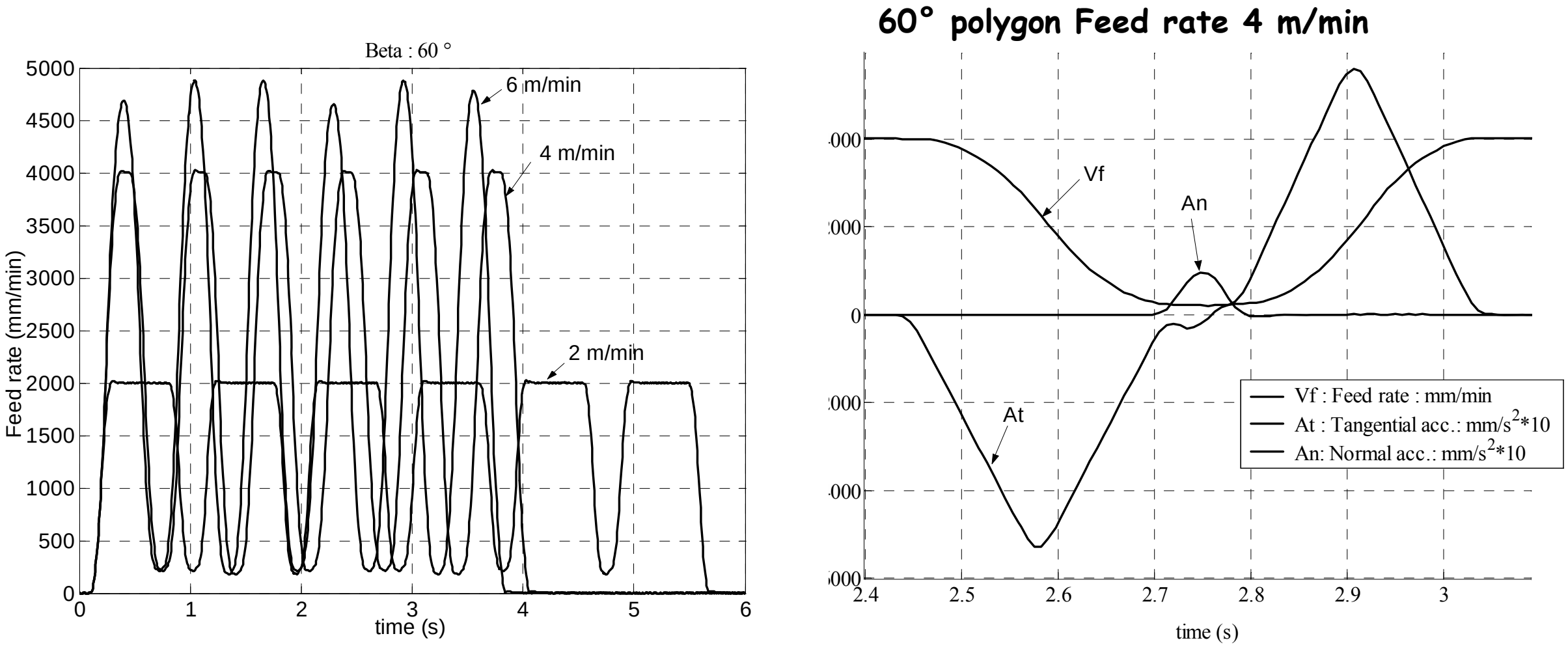

**a)- Feed rates record** **b)- Zoom in discontinuity**

*Figure 2. Records for a 60° polygon.*

Our experimental records show a significant reduction of the feed rate (Figure 2a) at the tangential discontinuities. The 840D controller of the MIKRON UCP 710 milling center integrates the hybrid solution presented above. The tool path is modified by the insertion of a curve which we represented by an arc of a circle and the reduction of the feed rate is not dependent on that programmed. The analysis of the kinematic behaviour shows that at the discontinuity the feed rate is constant and the tangential acceleration is null (Figure 2b). Moreover, it highlights the influence of the angle between two consecutive segments of the tool path (see Fig.1, angle $\alpha$) on the feed rate. Indeed, the feed rate decreases especially as the angle is small. This angle thus represents an indicator for performances evaluation.

*Curvature discontinuities*

Experiments are based on spiral tool paths programmed with the circular interpolation mode (G02 type). Different radii of curvature and feed rates are tested. Spirals are constructed with semi circles or quarter circles (example of semi circle spiral is shown in Figure 3). At each radius change 1, 2, 3, and 4 (Figure 3), a curvature discontinuity is created. The acceleration and feed rate records highlight the influence of curvature change. The decrease of the tangential feed rate is accompanied by variations of the normal acceleration.

As in the case of tangential discontinuities, the feed rate at the curvature discontinuity is independent of that programmed. Nevertheless its variation is linked to the curvature radius change. So, the curvature radius variation can also be retained as indicator.

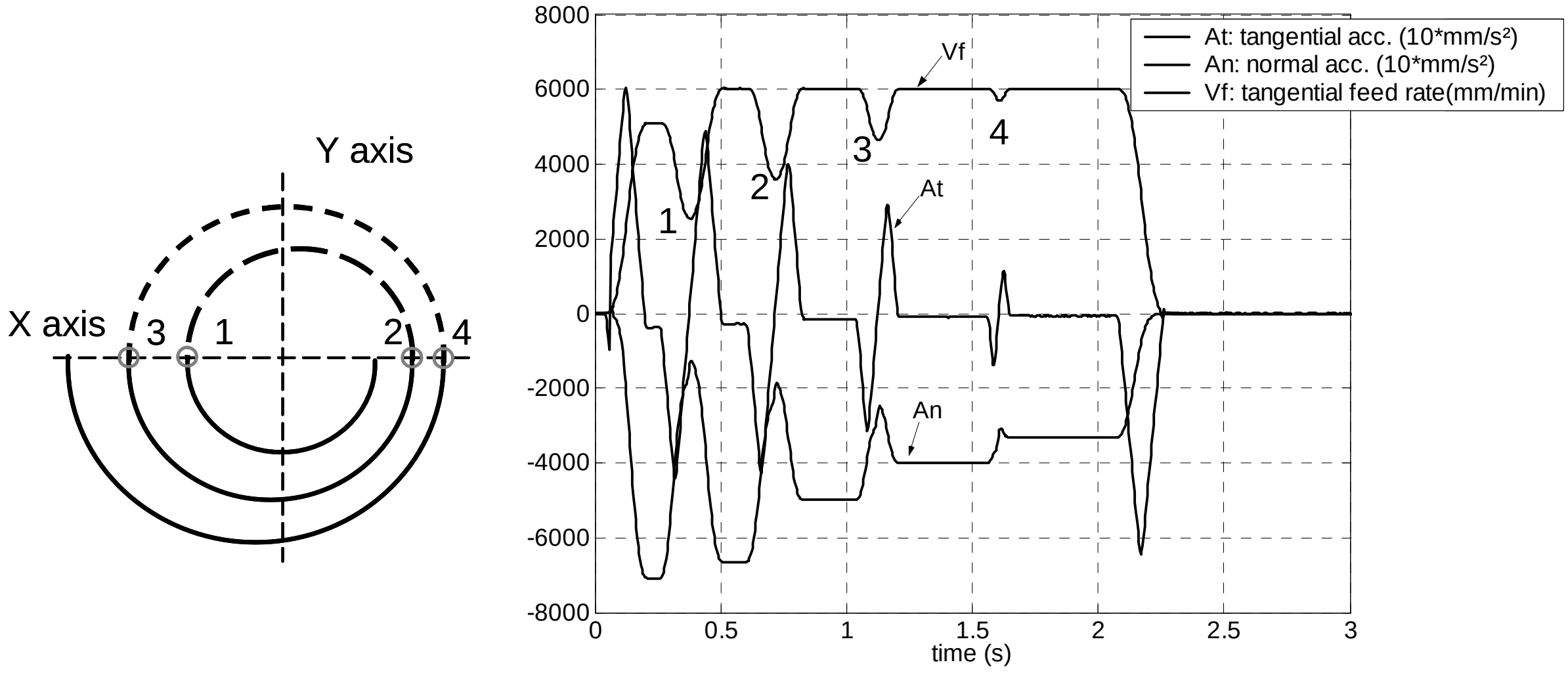

*Figure 3. Analysis of curvature discontinuities.*

## 2.2 NCU calculation capacity

Experiments are based on straight line tool path defined along each axis of the machine tool. Figure 4a shows an example of X axis. Between points A and B, the tool path is defined by successive points separated by a distance d. Between points B and C, only one segment of 30 mm is defined. The distance d takes different values as it is shown in Figure 4b. Tests are carried out with 10 m.min$^{-1}$ feed rate.

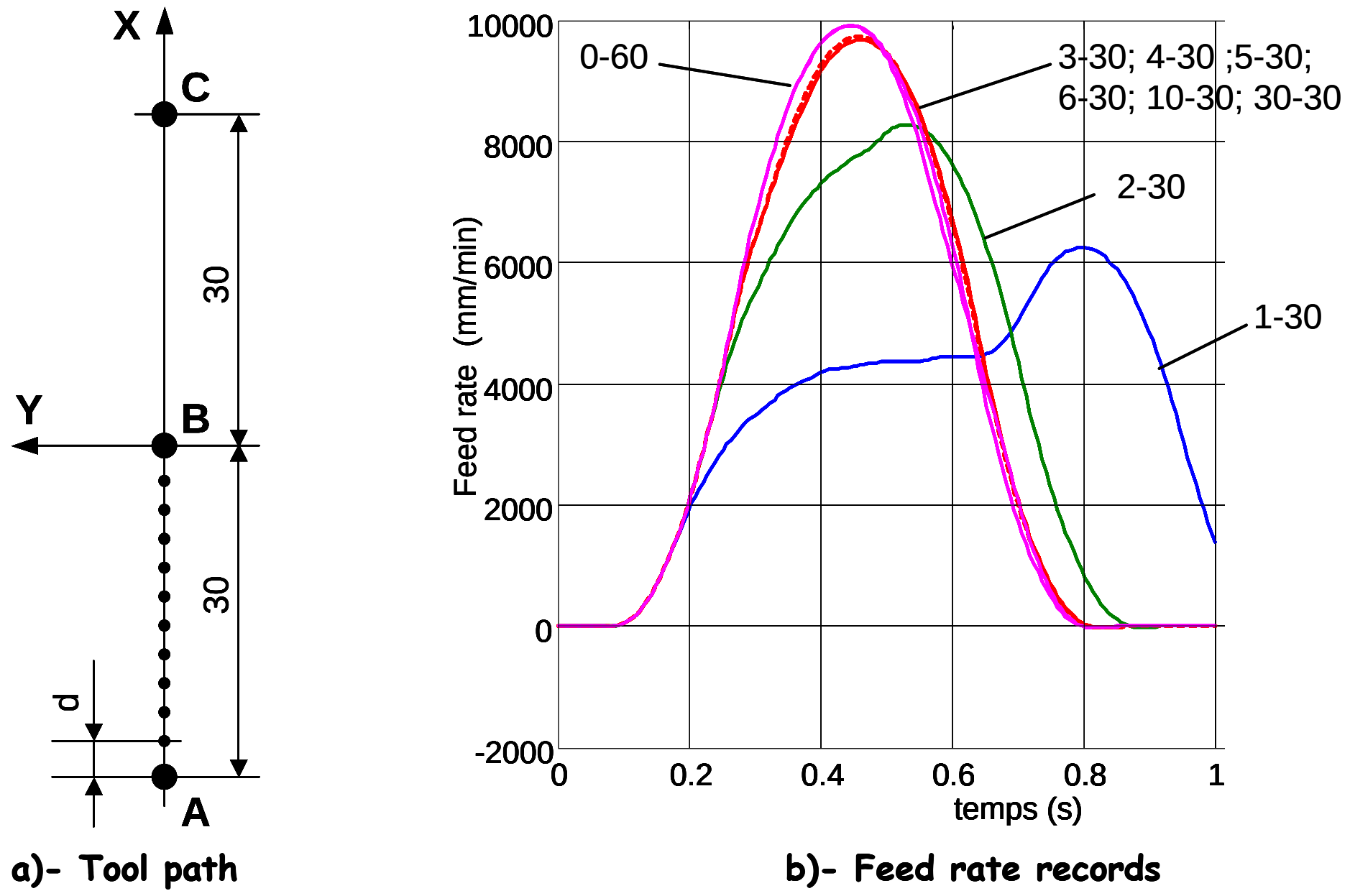

*Figure 4. Instantaneous feed rate surveys*

Over 3mm for the distance d, the programmed feed rate is reached. Under this value, the instantaneous feed rate decreases. For 1 mm distant points, the reduction of the instantaneous feed rate at point B is about 55%. The comparison between the two parts of the tool path (AB and BC) shows the insufficiency of the interpolation time to process blocks of 3 mm with a feed rate of 10 $m.min^{-1}$. Indeed the controller reduces, if necessary, the feed rate to allow the axis to cover the distance d with a time equal or more than the interpolation time. On the basis of these experiments, we choose the calculation capacity of the controller as indicator for the performance evaluation.

## 3 Performance viewer

The main objective of the performance viewer is to assist process development engineers during machining strategy elaboration. It is integrated after tool path computation with a CAM software. Indeed, to analyze the machining performance according to the chosen indicators, we use the machine tool technical constraints and the data resulting from the computation of tool paths. In the scope of our work the performance viewer is implemented in the ESPRIT® CAM software. ESPRIT® is programmed with objects defined in Visual Basic language. We implemented three main tasks (Figure 5). First, tool path objects are extracted (task A1). Then machine tool behaviour is evaluated according to the indicators (task A2). From the results of tasks A1 and A2, the machining performance is displayed in the ESPRIT® workspace (task A3).

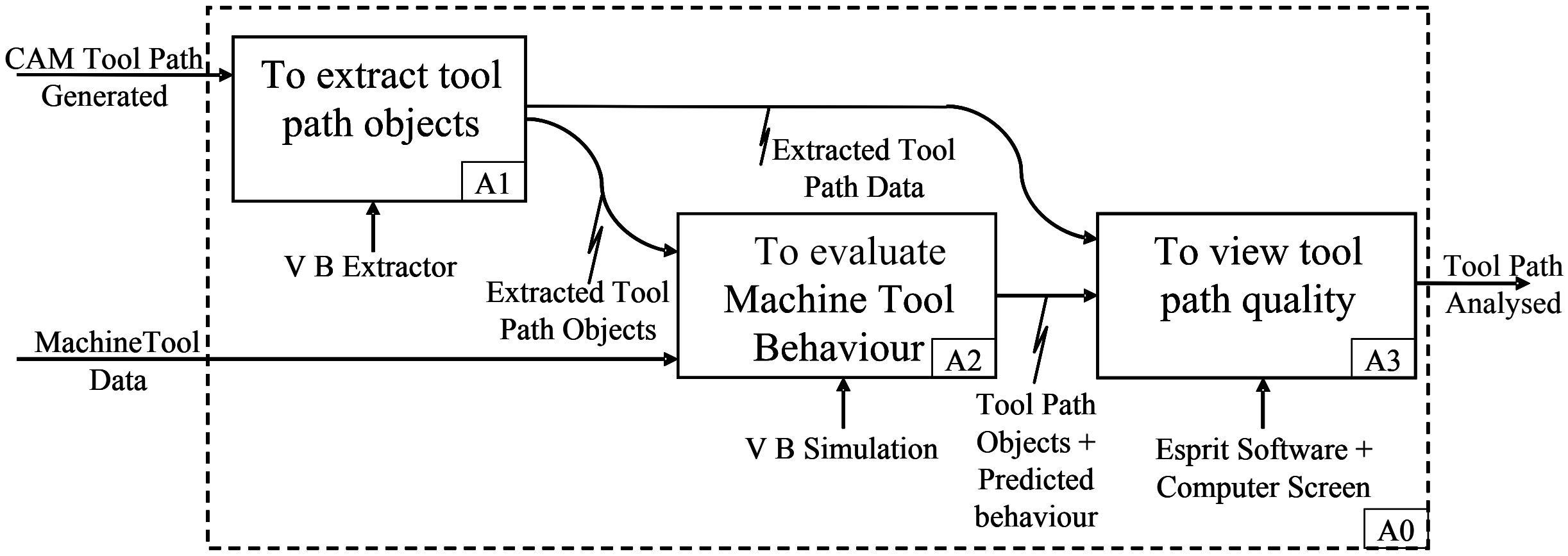

*Figure 5. Performance viewer : general structure*

### 3.1 Tool path objects extraction

Tool path extractor is divided in two sub-tasks which explore the tool path (object in VB language) selected in the workspace of ESPRIT®. Indeed, the tool path object is composed of tool path type, feed segments and rapid segments. The algorithm ~~we~~ developed analyses the tool path object in order to extract geometrical parameters (segment and arc) and machining data such as feed rates.

### 3.2 Machine tool behaviour evaluation

Machine tool evaluation behaviour integrates feed rate modelling. The models ~~we~~ used are based on two main works [11-12] and our experiments. As presented below, the data processed by the models are derived from the tool path object.

*Tangential discontinuity crossing model (model 1)*

In ~~our~~ this work, we consider that a circle arc is inserted at each tangential discontinuity, in the basis of the results of our experiments. So, the first step of machine tool behaviour modelling is to determine the arc radius. The value of the radius given by eq. (1) depends on the trajectory interpolation tolerance (TIT) (Figure 6). TIT identification on our machine tool is based on the test protocol proposed in [9] and our experiments.

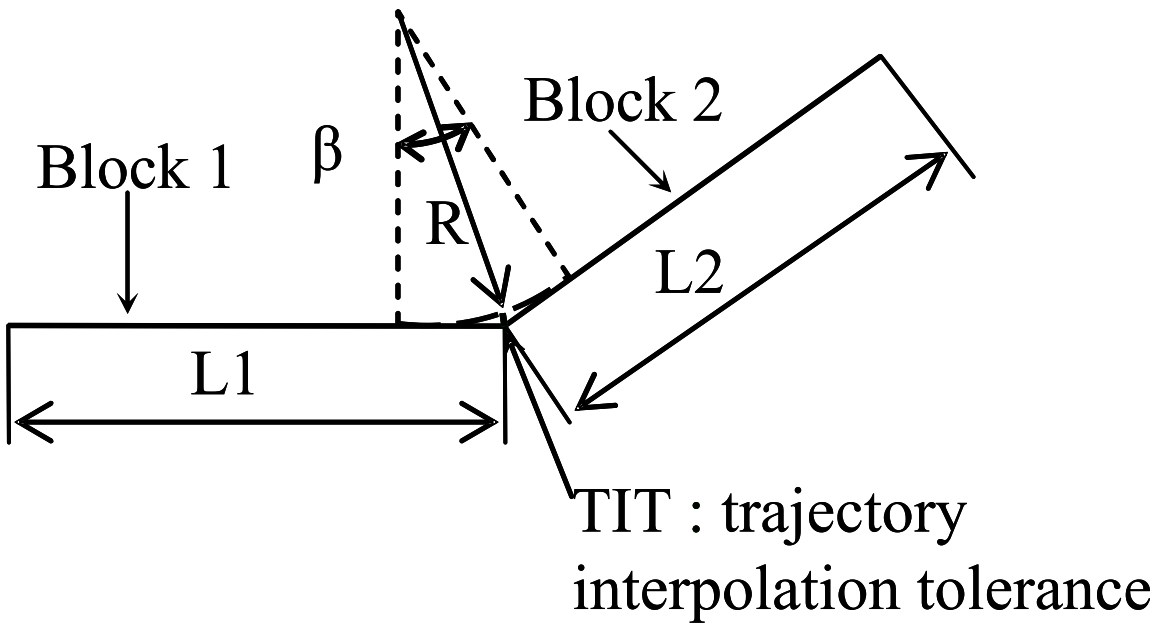

*Figure 6. Model 1: radius model calculation*

$$R = \min(TIT \times \frac{\cos(\beta/2)}{1-\cos(\beta/2)}; \frac{l}{2 \times \sin(\beta/2)} - TIT) \text{ with } l = \min(l1; l2) \quad (1)$$

The minimal feed rate is reached when the tangential acceleration is null. This phenomenon has been explained in literature and highlighted in our experiments (see Figure 2).

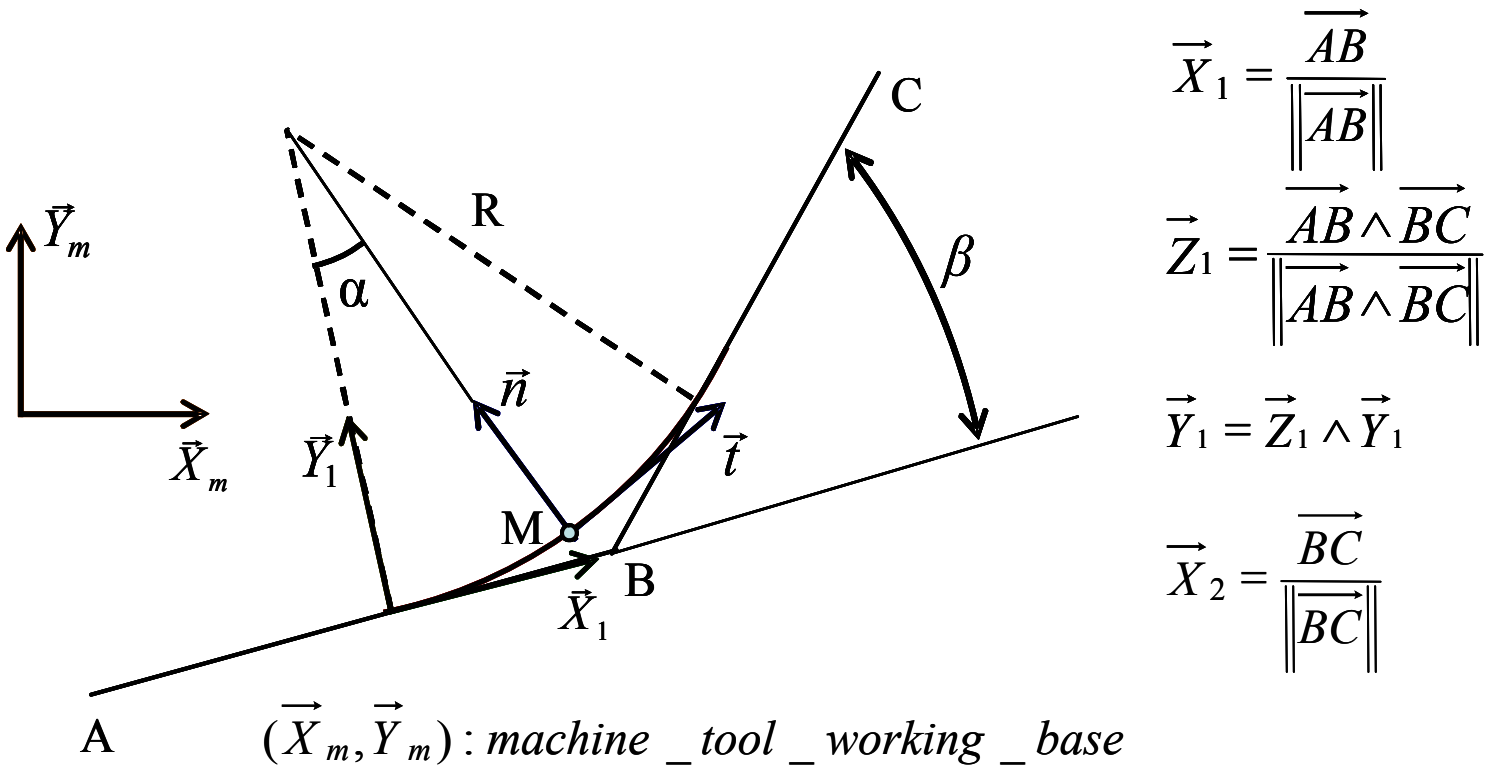

*Figure 7. Crossing arc radius inserted in XY plane*

In Figure 7, we consider that the minimal feed rate ($V_f$) is reached at the point M on the arc of a circle. So, we can deduce that at the M point the feed rate is constant and the normal acceleration can be calculated as follow:

$$\vec{A} = a.\vec{n} = \frac{V_f^{\,2}}{R}.\vec{n} \quad (2)$$

The jerk value can also be calculated as follow:

$$\vec{J} = \frac{d\vec{A}}{dt} = \frac{da}{dt}\vec{n} + a\frac{d\vec{n}}{dt} = \frac{V_f^{\,2}}{R}.\frac{V_f}{R}.\vec{t} = \frac{V_f^{\,3}}{R^2}.\vec{t} \quad (3)$$

According to eq. (2) and eq. (3), the feed rate is limited by the normal acceleration or the tangential jerk. The feed rate limited by the normal acceleration is deduced from eq. (2). Its value is given by:

$$V_{f_a} = \sqrt{a \times R} \tag{4}$$

Feed rate limited by the tangential jerk is deduced from eq. (3):

$$V_{f_jerk} = \sqrt[1/3]{Jerk \times R^2} \tag{5}$$

The feed rate can be also limited by the maximal tangential feed rate ($V_{max}$) according to the machine tool feed rate capacity. By taking into account the machine tool kinematic characteristics, we can deduce the feed rate limitation:

$$V_f = \min(V_{max}; V_{f_a}; V_{f_jerk}) \tag{6}$$

The kinematic characteristics (maximum feed rate, maximum acceleration and maximum jerk) given by machine tool manufacturers or evaluated are different for each axis. So, the tangential feed rate, the normal acceleration and the tangential jerk defined in the model 1 must be computed according to each axis capacity. We developed an algorithm to compute the values of these kinematic characteristics.

For instance in the case describes in Figure 7, the minimum feed rate is reached at the M point. The feed rate can be limited by the X axis or the Y axis capacity. To present our algorithm we consider that the limitation is introduced by the jerk of the X axis. This limitation is reached if:

$$\vec{t}(\alpha).\vec{Y}_m = 0 \Leftrightarrow \cos(\alpha) \times \vec{X}_1.\vec{Y}_m + \sin(\alpha) \times \vec{Y}_1.\vec{Y}_m = 0 \tag{7}$$

From eq. (7) the restrictive angle ($\alpha_{J_X}$) associated to the jerk limitation for the X axis can be deduced. Then the tangential jerk at point M is computed as follow:

If $\alpha_{J_X} \in [0, \beta]$ then $Jerk = J_X$ where $J_X$ is the maximal jerk of the X axis. (8)

Else if $\vec{t}(0).\vec{Y}_m > \vec{t}(\alpha).\vec{Y}_m$ then $Jerk = \dfrac{J_X}{\left|\vec{X}_1.\vec{Y}_m\right|}$ (9)

Else $Jerk = \dfrac{J_X}{\left|\vec{X}_2.\vec{Y}_m\right|}$ (10)

The same computation process can be applied for the maximal feed rate and acceleration for the Y axis in our example. Finally, the Jerk values given by eq. (8), eq. (9) or eq. (10) for the two axes X and Y are compared. The value of the tangential jerk used in model 1 eq. (5) is the minimal one. The same reasoning is made to deduce the tangential feed rate and the normal acceleration defined in model 1.

*Curvature discontinuity crossing model (model 2)*

To calculate the feed rate when crossing a discontinuity between a segment and an arc, we use the corner radius model developed in [12]. As it is shown in Figure 3, the tangential acceleration is null at the curvature discontinuity. The normal acceleration also changes at the curvature discontinuity and its value can be written in a Frenet frame ($\vec{t}, \vec{n}$) as follow:

$$\left|\vec{A}(t + \delta t/2) \bullet \vec{n} - \vec{A}(t - \delta t/2) \bullet \vec{n}\right| = Jerk \times \delta t \tag{11}$$

δt is an elementary time variation necessary to pass the discontinuity. This elementary time has been evaluated for our machine tool during the works carried out by Pateloup [13]. It has also been identified during our experiments.

When crossing the discontinuity the jerk is tangential and the feed rate $V_p$ is constant. So, only the normal acceleration can be calculated with the programmed arc radius R as follow:

$$\left|\vec{A}(t-\delta t/2)\bullet\vec{n}\right| = 0 \text{ and } \left|\vec{A}(t+\delta t/2)\bullet\vec{n}\right| = \frac{V_p^{\,2}}{R} \tag{12}$$

The feed rate of the model 2 given by eq. (13) can be deduced from eq. (11) and eq. (12):

$$V_p = \sqrt{Jerk \times \delta t \times R} \tag{13}$$

A curvature discontinuity model for two consecutive arcs was developed by Pateloup [13]. Its observations about the tangential acceleration and the normal acceleration correspond to those we did for an arc and a segment. Thus, by applying the reasoning of Pateloup, the feed rate at the discontinuity can be calculated as follows:

$$V_{fr} = \sqrt{\frac{R_{r1} \times R_{r2} \times Jerk \times \delta t}{\left|R_{r1} - R_{r2}\right|}} \tag{14}$$

Where jerk is the tangential one, Rr1 and Rr2 are respectively the radius of curvature before and after the discontinuity, δt is the elementary time variation necessary to pass the discontinuity.

For each model, the tangential jerk is computed according to its capacity. In the example shown in Figure 8, the discontinuity appears at the point B. The tangential jerk can be calculated as follow:

$$Jerk = \min\left(\frac{J_X}{\left|\cos(\alpha)\right|};\frac{J_Y}{\left|\sin(\alpha)\right|}\right) \tag{15}$$

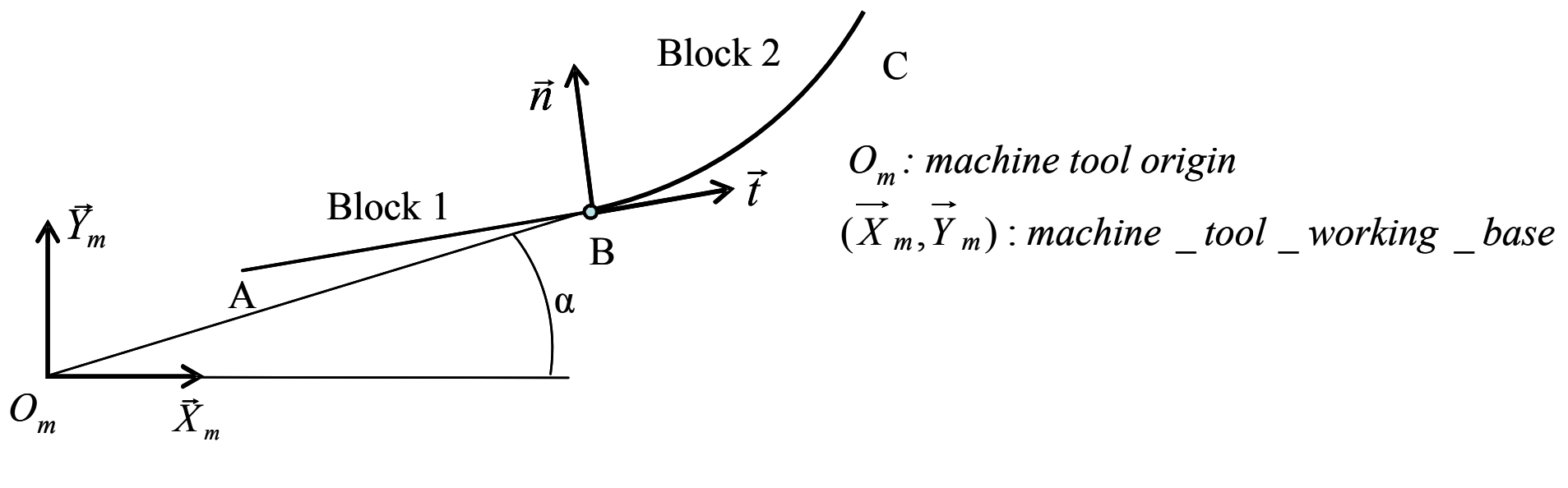

*Figure 8. Crossing curvature discontinuity in XY plane*

*NCU capacity model (model 3)*

As we observed in our experiments (Figure 4) the controller calculation capacity is linked to the minimal distance L between to consecutive points on the tool path. For a given programmed feed rate, the distance L can be evaluated from the interpolation time $t_{int}$ of the NCU as follow:

$$L = V_p \times t_{int} \tag{16}$$

The interpolation time is difficult to evaluate with experiments [9]. To calculate the distance L, we use the value of $t_{int}$ given by Siemens for the 840D controller ($t_{int} = 12 \times 10^{-3}$ s)

### 3.3 Tool path quality display

The tool path quality evaluation is based on the comparison of the programmed feed rate and the predicted feed rate. According to a colour code, this comparison is displayed on the ESPRIT® workspace. The three indicators retained are used for the evaluation of the tool path: NCU calculation capacity and discontinuity crossing (tangential and curvature discontinuities). Figure 9 shows the general structure of the quality display module.

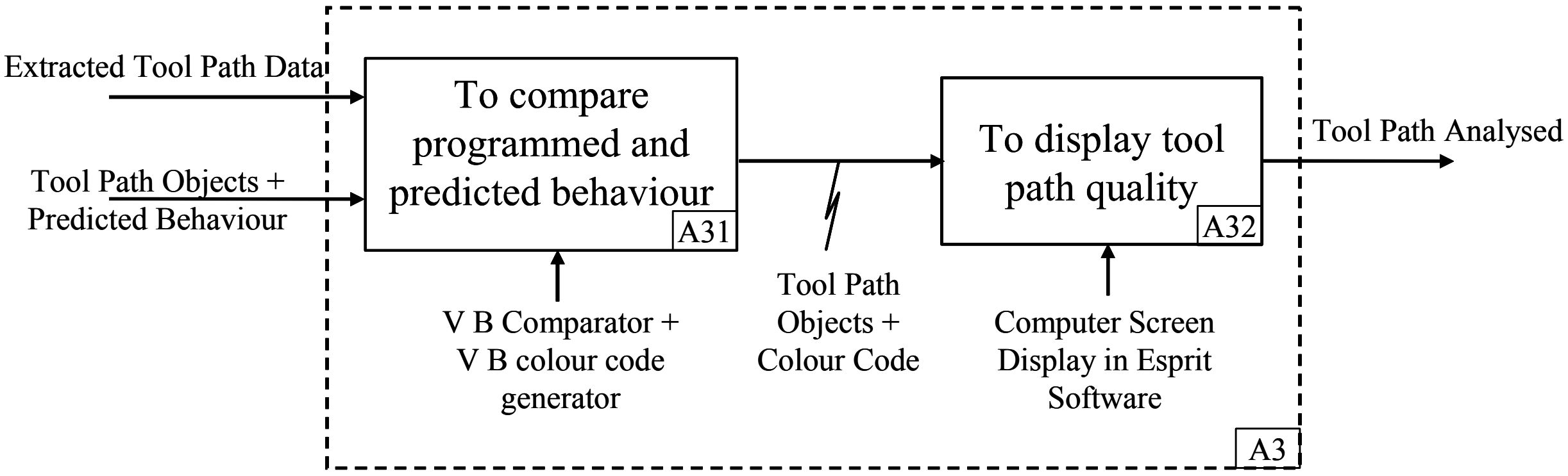

*Figure 9. Tool path quality display structure*

*NCU calculation display*

Each segments or arcs length is read and according to eq. (16) the length is compared with the minimal distance L which corresponds to the programmed feed rate. Then each segments and arcs are classified according to its length. Figure 10 shows an example of the evaluation carried out during a Z-level spiral finishing tool path.

The machining strategy performance analysis is made according to the capacity of the MIKRON UCP 710 milling center. Two main observations can be made from this analysis. First, several small segments or arcs have been generated with the chosen machining strategy. This observation is illustrated by the high number of portions of tool paths beneath the minimal distance L. Second, some areas generated more small segments and arcs than others. These areas correspond either to features having small dimensions, or with the inadequacy of the selected machining strategy. From these observations, process development engineer can locally modify the tool path or globally change the machining strategy.

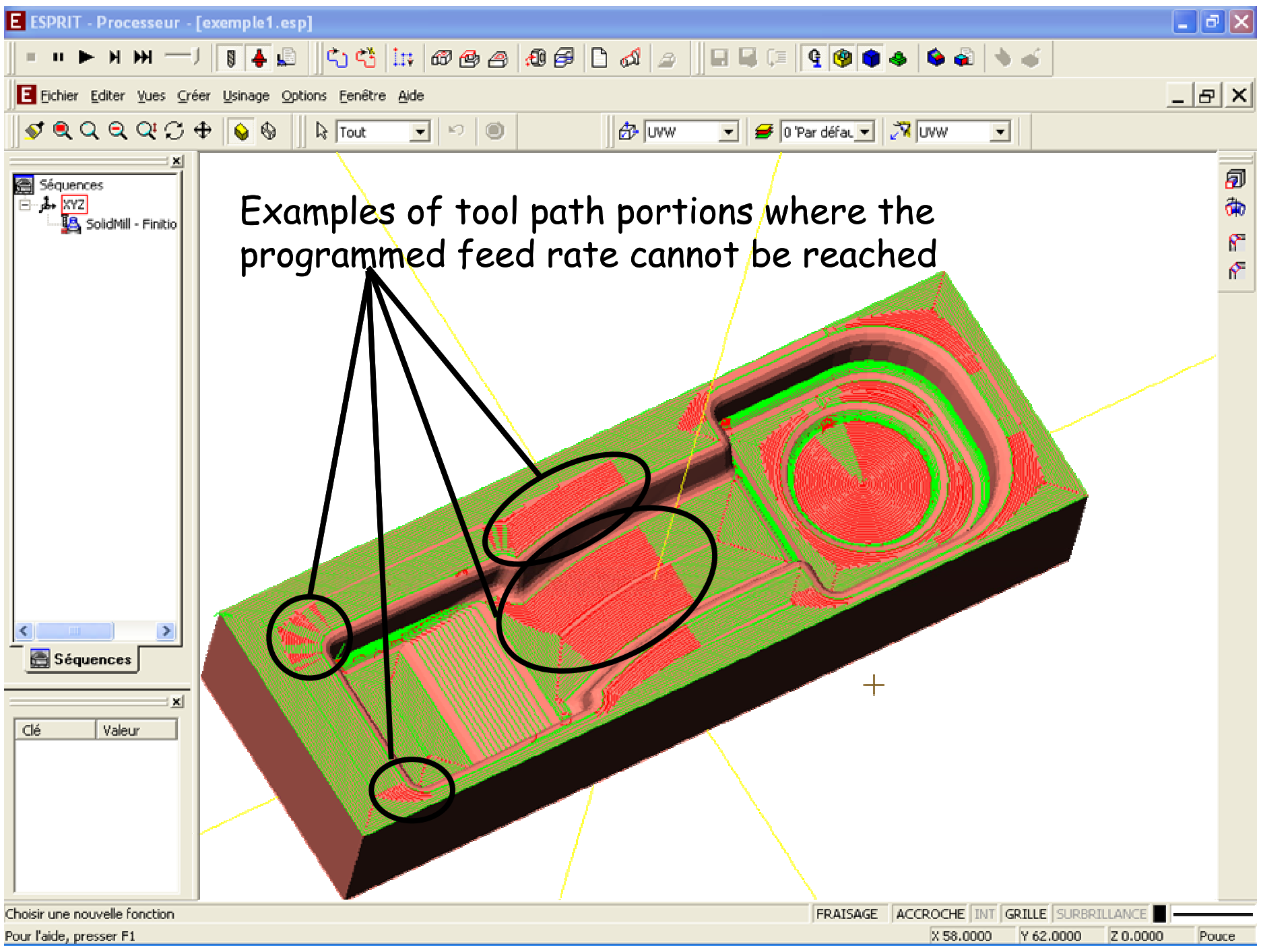

*Figure 10. NCU capacity Display*

*Discontinuity crossing display*

Figure 11 shows the machine tool behaviour visualization at discontinuities. The predicted feed rates computed according to models 1 and 2 are compared with the programmed feed rates. Each discontinuity crossing point is classified according to the programmed feed rate reduction percentage.

The visualization underlines the areas already identified during controller calculation capacity (red color in Figure 10). Indeed, the areas in which the length of the segments and the arcs are lower than the minimal length L correspond in general to a reduction of more than 90% of the programmed feed rate. This observation highlights the complementarity between the two types of visualization and indicators relevance. It also makes it possible to retain the inadequacy of the machining strategy for these areas.

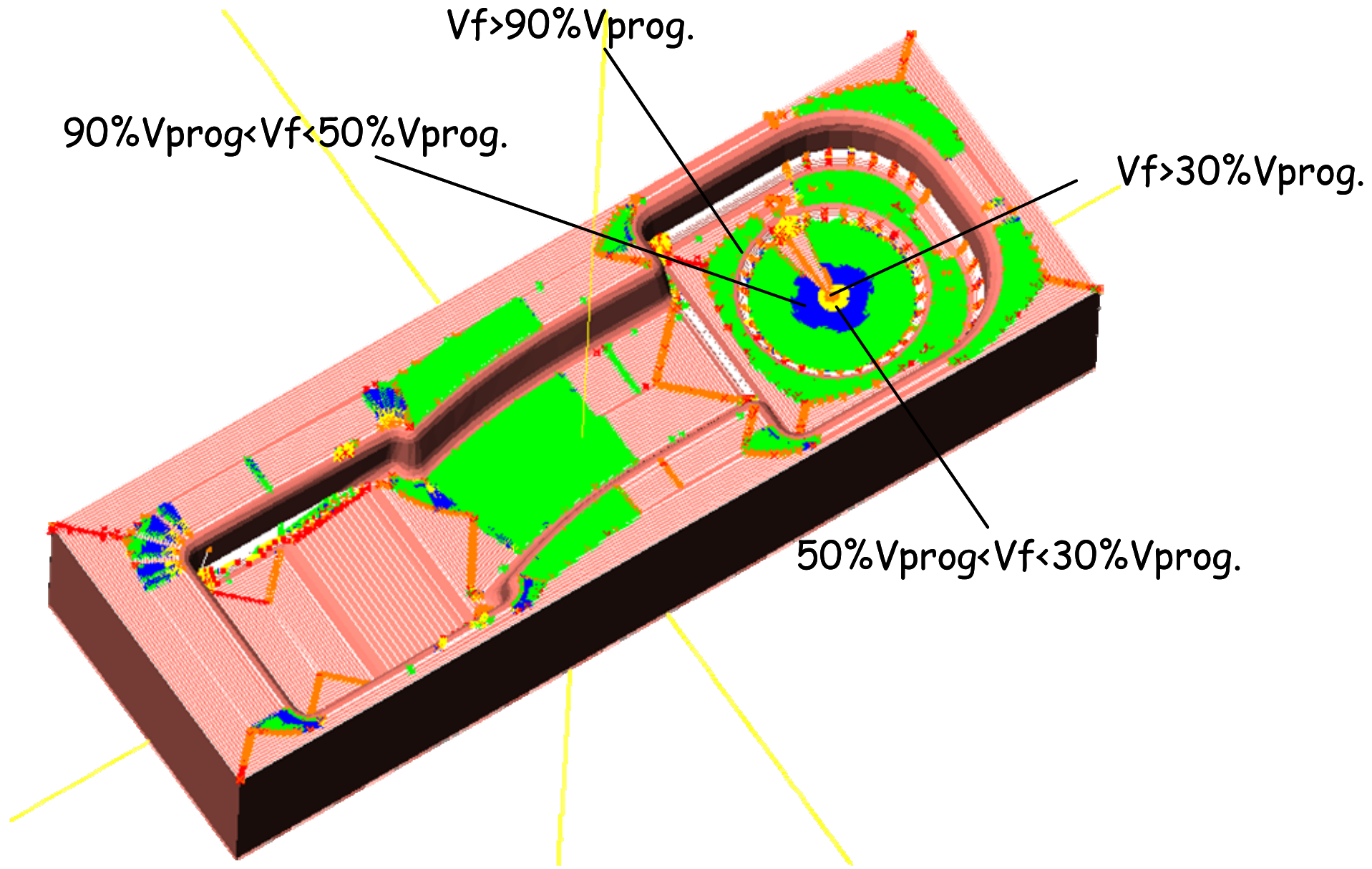

*Figure 11. Crossing discontinuity display*

## 4 Conclusion

In this paper we propose a performance viewer developed in ESPRIT® CAM Software. Three main modules have been developed: tool path object extraction, machine tool behaviour evaluation and tool path quality visualization. As the performance viewer is integrated in the ESPRIT® workspace, the development of the modules is carried out using the Visual Basic language. The machine tool behaviour evaluation is based on the prediction of the feed rate according to geometrical elements (segments and arcs) found in the tool path. The indicators identified for this evaluation are controller calculation capacity and two machine tool kinematical behaviours: tangential and curvature discontinuities. These indicators have been modelled and experiments have been performed to determine some parameters of the models.

Our work is focused on three axis tool path often used in mould and die machining. In this context, linear and circular interpolated trajectories are analysed with our performance viewer. Thanks to this viewer, die and mould makers are able to analyse machining strategies and to evaluate the tool path quality according to the machine tool capacities.

The future applications of the work presented in this paper are directed towards the decomposition of topology of moulds and dies in order to improve the quality of machining. This decomposition is an alternative to the machining features recognition which is very difficult to be implemented in the case of complex parts machining.